\def\year{2017}\relax
\documentclass[letterpaper]{article} 
\usepackage{aaai17}  
\usepackage{times}  
\usepackage{helvet}  
\usepackage{courier}  
\usepackage{url}  
\usepackage{graphicx}  
\usepackage{listings}
\begin{document}
%
\title{A Vision For Continuous Automated Game Design}
\author{Michael Cook\\
Games Academy\\
Falmouth University\\
mike@gamesbyangelina.org
}
\maketitle
\begin{abstract}
ANGELINA is an automated game design system which has previously been built as a single software block which designs games from start to finish. In this paper we outline a roadmap for the development of a new version of ANGELINA, designed to iterate on games in different ways to produce a continuous creative process that will improve the quality of its work, but more importantly improve the perception of the software as being an independently creative piece of software. We provide an initial report of the system's structure here as well as results from the first working module of the system.
\end{abstract}

\section{Introduction}
Procedural generation continues to flourish both in the world of game development and academic research, and we are past the days of considering it only as a means of saving time or money. Additionally, the role of generative software in games can extend beyond simply generating content for one particular game, and we are seeing more and more interest now in building automated game designers -- software systems that generate games in their entirety, rather than creating content to be used in a larger, otherwise static creation.

Automated game design is a particularly important point of crossover between game AI research and \textit{computational creativity}, a subfield of AI concerned with software that can assist humans in being creative, or act creatively on its own \cite{frontiers}. Computational creativity contends that the \textit{perception} of a piece of software is as important as any quantifiable properties the software or its output might have \cite{perception}. In other words, how we feel about an automated game designer is just as important as whether it produces good games.

\begin{figure}[h]
\centering
\includegraphics[width=0.8\columnwidth]{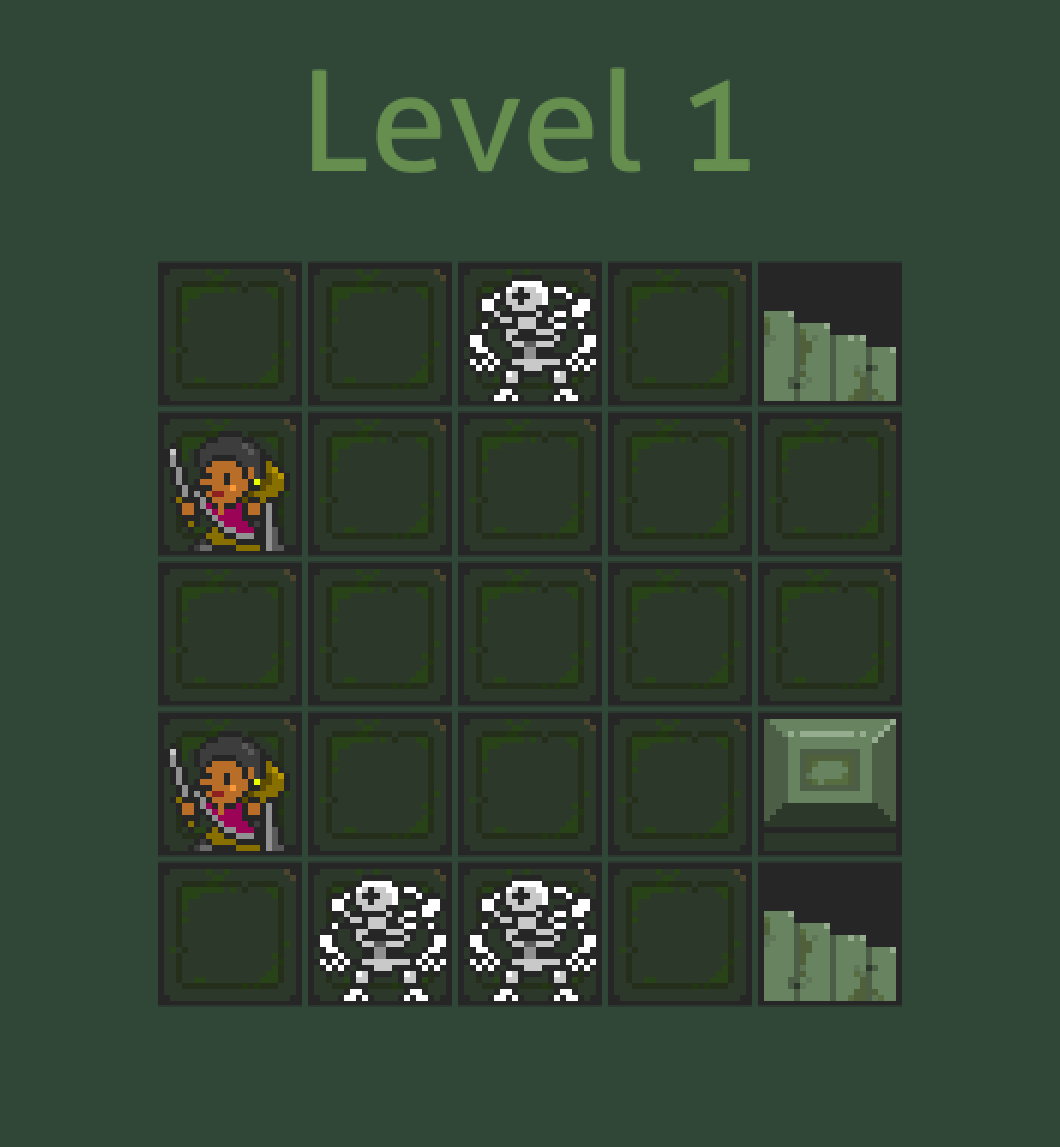}
\caption{A screenshot of a level from \textit{Before Venturing Forth}. Art by Oryx.}
\end{figure}

This means rethinking a lot of the ideas we might have about evaluation, inherited from areas like procedural generation research. A procedural generation usually is judged by how good the content it generates is. But for automated game design we must also consider how our system presents itself, how it responds to interrogation (if such a thing is even possible), how it grows over time and demonstrates this to the people who engage with it.

In this paper I outline a roadmap for a new version of ANGELINA, an automated game designer. 
Partly it lays out future research we I intend to carry out, but it's also a broader statement about how automated game design can be done, and how perhaps we can grow the systems we build in this area beyond simple one-shot processes that are given some data and produce a game at the end. I'm going to show how I hope to build a creative profile for ANGELINA that can be followed and engaged with, and how this is more than cosmetic but should also meaningfully impact the games the system makes, too.

The remainder of the paper is as follows: in \textit{Related Work} I cover some existing work in automated game design and talk about how many of these systems are structured; in \textit{ANGELINA} I outline the current state of ANGELINA and its long-term goals. This section also describes the motivation for continuous automated game design in creative terms, and discusses how the system and its description language has been designed to facilitate both high-level abstract design and low-level experimentation and invention. Finally, in \textit{Future Work} I discuss the long-term goals for automated game design, and why I think continuous approaches are important. 

Automated game design is hard, and we are unlikely to see it have huge impact on mainstream game development any time soon. Nevertheless, it is a rich and exciting area to research in, and there are many unanswered questions and early ground still to be covered. It has the potential to shed a light on the broader field of generative software too, by providing opportunity to reflect on development and design processes, and to understand how people respond to game content created by machines. It is more than another subfield of procedural generation: automated game design lets us ask important questions about the involvement of AI in creative practices, in a medium that is inherently digital, interactive and fluid. This paper is a short roadmap for some important goals I see on the horizon, but more than anything else we need more people and perspectives (particularly outside of hard AI) on this multifaceted and complex problem.

\section{Related Work}
Automated game design is often conflated with the notion of ruleset design. One explanation for this is the bias in games research towards a particular `classical' concept of what a game is. Early research in AI focused on abstract games that are almost entirely described by a set of rules and nothing else (Checkers, Chess and Go being the best examples here). As games research moved more into digital games, classic arcade stereotypes replaced this notion; games with very strong notions of winning, losing, scoring. We can see this trend continuing to the modern day in the design and influences of the Video Game Description Language, VGDL \cite{schaul2013pyvgdl}.

As a result of this conflation, a lot of automated game design works pays homage to work on physical games, like the invention of Chess variants by Pell in \cite{metagame}. This pioneering work was extremely safe in many ways: starting off from a solid foundation of human design, searching the nearby ruleset space with no other design elements to consider. Nevertheless, it represents an important step in appreciating that all game content can be generated or changed by a computer, not merely surface-level consumable content.

The line of work on ruleset invention carries through to work on systems like Ludi, by Cameron Browne, which invented a broader variety of abstract games played with counter on grid boards \cite{brownethesis}. Ludi is a compelling example of impacting a creative space -- one of its game designs was published successfully, and has a good ranking on BoardGameGeek's comprehensive and vast index of games. At the time of writing it is ranked \#3,637 (out of over 90,000 games) and ranked \#104 in Abstract Games. Like the Chess variants, Ludi works only on rulesets - it doesn't distinguish between different design tasks, or consider any artistic qualities in its work. This doesn't in any way diminish its significant achievements - we mention it here only to underline the many other tasks we see in the field of automated game design distinct from designing rulesets.

In terms of digital games, the Game-O-Matic is a vital part of automated game design history \cite{gom}. Ostensibly developed as a mixed-initiative tool for journalists to rapidly create newsgames, the Game-O-Matic is effectively an entirely autonomous game designer, with a very broad understanding of game mechanics and also how to convey messages through the combination of game rules. This meaning-driven game design is reinforced by several mechanisms that allow the Game-O-Matic to retrieve artistic assets to reinforce the game's systems visually. Along similar lines, work by Nelson and Mateas in \cite{meaning} also shows a concerted effort to build a system which can combine systems, meaning and visuals to convey something through an interactive experience. 

There are also a range of tools that, while not built as autonomous designers, are sufficiently  close in nature that they might only be a few small changes from being such. Tools like \textit{Tanagra} \cite{tanagra} use a lot of automation and self-analysis to provide support to a human user, or the \textit{Sentient Sketchbook} \cite{sentient} which amplifies smaller creative inputs with embellishments and analysis, and \textit{Ropossum} \cite{ropossum}, a similar tool targeting physics-based mobile games like \textit{Cut The Rope}. These tools are performing a lot of the design work going on in their domain, even if the project's intention is not to fully automate the task. Often, the need to support humans in any part of a creative tasks means the system must, by definition, be able to perform every part of the task on its own. Such systems are usually only lacking in some higher-level control and production, and are otherwise very close to automated game design tools.

\section{ANGELINA}
Between 2010 and 2016, several versions of ANGELINA were developed. Most of them followed a similar modular development structure which broke the process of designing games into several processes which ran concurrently \cite{phdthesis}. For example, the third iteration of ANGELINA --  ANGELINA$^3$ -- had systems for designing levels, tuning the design of in-game upgrades, and laying out art and sound assets for maximum impact \cite{angelina-aiide}. All ANGELINA versions were designed as \textit{single-run productions}, meaning that the system was turned on, given input, worked for a time, produced a game, and then stopped.

Between runs, very little information was retained. The amount of data ANGELINA retained did increase over time, but even later versions only kept information relating to cultural knowledge and art assets. For example, ANGELINA$^5$ built a list of cultural information and affective texture labels (a texture of grass gained labels like \textit{bright}, \textit{nature}, \textit{fresh} over time) so that it could make better design decisions when it designed games in the future.

A key goal with ANGELINA was presenting the system as a plausible and respectable creative entity. That is, a system which could engage with its peers in a creative community, produce work that people engaged with and were interested in; a system which could show growth, justify its decisions and explain its process. Most of the later versions of ANGELINA wrote commentaries to explain their game designs, and ANGELINA entered game jams and exhibited in galleries. Nevertheless, a persistent issue was that ANGELINA itself, the system which produced games, existed behind a curtain and was rarely encountered by the people who played its output. ANGELINA's creativity could only be experienced once-removed, through the things it produced - either games or commentaries.

Exposing ANGELINA's creative process has always been \textit{possible}, in the most literal sense that someone could watch it produce a game over several hours. But in terms of presenting a creative process that was worth following or studying, ANGELINA has been lacking on a fundamental level. There is very little long-term growth to observe, and little change from the first game it designed to the hundredth. The creative process is also simplistic and discretised, due to the single-run nature of the system. Each time the system begins with a blank slate and doesn't stop until it has gone through the same rote set of steps. In reality, creative processes are fluid, they start, stop, restart and change course; they might even evolve into other projects entirely. People follow the creation of videogames because they are interested in the twists and turns, the growth of ideas, the surges of inspiration and the slow reveal of a project -- and an artist -- growing from start to finish over a long time.

To solve some of these accessibility and visibility problems, we are planning a new structure for ANGELINA which breaks down game design into a collection of separate activities which are fluidly moved between, with each activity forming part of a \textit{continuous game design process} that has no particular start or end. These tasks inform one another - some tasks create knowledge or concepts which are reused elsewhere, and designing a full game typically requires some movement between all of these activities. This means that the system can be seen to be changing over time, developing new ideas and new knowledge. It also means that the creation of a game is not the beginning or the end of the process, but a side effect of the system's creative activity.

\section{Continuous Design}
In this section we discuss the notion of Continuous Design in more detail, and outline the intended structure of this new ANGELINA, the format we're using to define games for this system, and why it matters. Continuous automated game design consists of a set of creative activities which are isolated from one another but contribute to a larger body of knowledge and ideas that are shared by all the creative states the system can be in.

Many of these creative activities can be \textit{nonlinearly composed} with one another, meaning there is not a strict order in which activities must be undertaken. For example, a system might move from designing levels for a specific game, to prototyping a new mechanic idea that may not be relevant to the current game being designed. It might note down an idea for a game theme that it returns to many weeks later, or discover an idea that fixes an abandoned half-finished idea it gave up working on days ago.

Free movement between these activities is important in demonstrating that the system has autonomy, and is not simply guided towards the goal of spewing out creative artefacts. The system might only choose to produce games it finds sufficiently interesting. By making the creative process a subject of interest we no longer have to worry about only evaluating our systems on the basis of the games they output. The mantra of prioritising \textit{process} over \textit{product} has been prominent in Computational Creativity literature for many years. Our attempt to follow this by producing commentaries that explain generated games is useful, but it still only allows engagement with products. By creating a system that is designed to be watched and followed, we allow the system to demonstrate its creative skill and appreciation regardless of whether or not it ever produces anything. We believe this could be a major step forward for the perception of creative software, and specifically within game design.

In the remainder of this section we give some background on this new version of ANGELINA by explaining our game description language and its role within the creative system. In the next section we'll describe our current plan for ANGELINA's internal structure, expressed through creative activities.

\lstdefinestyle{customc}{
  belowcaptionskip=1\baselineskip,
  breaklines=true,
  frame=L,
  xleftmargin=\parindent,
  language=C,
  showstringspaces=false,
  basicstyle=\footnotesize\ttfamily,
  keywordstyle=\bfseries\color{green!40!black},
  commentstyle=\itshape\color{purple!40!black},
  identifierstyle=\color{blue},
  stringstyle=\color{orange},
}
\lstset{basicstyle=\footnotesize\ttfamily,}

\begin{figure}[t]
\begin{lstlisting}
    "gamename": "Before Venturing Forth",
    "numplayers" : 1,
    "floor": "dungeonfloor",
    "music": "ominous",
    "color_accent": [0.4, 0.56, 0.31],
    "color_body": [0.19, 0.28, 0.22],
    "variables" : [
        {
            "name": "score",
            "onscreen": "Score",
            "startvalue": 0
        }
    ],
    "pieces" : [
        {
            "name": "playerpiece",
            "layer": 5,
            "sprite": "fighter",
            "animated": true,
            "flips": true
        },
        // ... edited for length
        {
            "name": "enemy",
            "layer": 5,
            "sprite": "golem",
            "animated": true,
            "flips": true
        }
    ],
\end{lstlisting}
\caption{The preamble from the game description for \textit{Before Venturing Forth}.}
\label{preamble}
\end{figure}

\begin{figure}[t]
\begin{lstlisting}
{
    "trigger": "OVERLAP playerpiece enemy",
    "code": [
        "DESTROY $2",
        "SFX punch",
        "SCORE 1"
    ]
},
\end{lstlisting}
\caption{A rule excerpt from the definition file for \textit{Before Venturing Forth}.}
\label{languagesample}
\end{figure}

\begin{figure}[t]
\begin{lstlisting}
{
        "type": "raw",
        "width": "5",
        "height": "5",
        "data": [0,4,4,0,3,
                 1,0,0,0,2,
                 0,0,0,0,0,
                 1,0,0,0,0,
                 0,0,4,0,3]
},
\end{lstlisting}
\caption{A level from the definition file for \textit{Before Venturing Forth}.}
\label{levelsample}
\end{figure}

\subsection{Game Representation In ANGELINA}
Historically our approach to defining games within ANGELINA has drifted between two extremes: very high-level description languages, and very low-level code manipulation. In the former case, systems like the very first ANGELINA were provided with abstract chunks of game design knowledge that could be reassembled into games by combining them with other chunks of knowledge. For example, the AI behaviour \textit{NPC chases Player} is an atomic concept in the first ANGELINA. Although it can intelligently evaluate the use of this concept in a game, it doesn't understand what the concept does, it can't change it to make it its own, and it can't explore adjacent ideas or innovate around this concept. This not only cuts down on what we can expect from the system's output, but it also harms the perception of the system - people regularly criticised ANGELINA for inheriting a lot of knowledge from its creator.

At the other extreme, we have experimented with direct manipulation of game engine code. In \cite{mechanicminer} we built a system that used metaprogramming techniques to search for game variables and write one-line operations that changed the variable in some way when a button was pressed. The system then experimented with these generated controls to see what gameplay affordances it offered, if any. While this experiment was successful in discovering new ideas, the painstaking level of detail it worked at meant its overall design process was much slower, and rapid prototyping or exploration of ideas was much harder to do. Both of these approaches had their strengths, but by only using one in a single system it was hard to avoid their weaknesses.

In this new continuous approach to game design we wish to build a game representation format that is suitable for design work across multiple creative activities -- in other words, a representation that supports both high- and low-level work. Our representation format uses a JSON structure to define a game, with different definition phases inspired by PuzzleScript \cite{puzzlescript}. Figure \ref{preamble} shows the preamble for a game called \textit{Before Venturing Forth}, with areas for defining cosmetic settings, then a list of variables, then a list of game objects. Subsequent sections then refer back to these - Figure \ref{levelsample} shows a level design, which uses integer codes specified in the \verb|pieces| section to concisely arrange game objects in the world. Although the language is still changing as the system develops, we have put a sample game definition file online so the reader can see a full game defined\footnote{www.gamesbyangelina.org/exag/example.json}. Note that this game is mostly designed by a human, although ANGELINA created level designs based on its interpretation of the mechanics. We include it here largely as a proof of concept for the language design.

The most important element of the design language is the way rules are defined. Figure \ref{languagesample} shows a rule definition from the linked example game. Rules have a head and a body -- the head is a \verb|trigger| that represents some condition that is checked for by the game; the body is a \verb|code| segment consisting of a series of operations that are applied in sequence. We have some simple syntax for passing parameters through triggers and into code. This example rule is triggered by pieces of two types overlapping with one another, which triggers a series of events to be executed in response. \verb|DESTROY $2| destroys the second game object referenced in the trigger (in this case, an \verb|enemy| piece), \verb|SFX punch| plays a named sound effect, and \verb|SCORE 1| adds 1 to the variable \verb|SCORE|.

This notion of a `rule' is fundamental to enabling ANGELINA to work across multiple different creative activities. As we will see in the following descriptions of creative activities, rules are designed to be useful at high and low levels of detail. When working with rulesets, ANGELINA might select game design patterns from a catalogue, with a pattern consisting of one or possibly multiple rules. ANGELINA can use these blindly without knowing what the rules contain, much like ANGELINA$^1$ did. Equally, in another creative activities, ANGELINA can create its own rule blocks by choosing triggers and events and combining them into a new block. One or more blocks might end up back in a catalogue of patterns, able to be used blindly in future creative activities. The creative work and knowledge discovery has already been `banked', meaning ANGELINA can use a rule or rules as a whole, knowing that it has previously confirmed the rule's usefulness or interestingness.

This allows us to find a compromise between the two extremes of ANGELINA's previous work. If we wish, we can work at low levels of detail, and painstakingly tune individual rules until they show promise. But equally, the system can work rapidly, prototyping games using existing rule knowledge and not worrying about fine-grained invention of mechanics. This representation of rules is flexible enough to be used in many different ways, and we eventually hope to have ANGELINA even invent new triggers and events, allowing it to dive even further into detail in its creative work.

\section{Creative Activities}
In this section we describe the three activity modules we are currently working on for the new version of ANGELINA. \textit{Level Design} is the most developed one currently, and \textit{Mechanic Invention} the least. We plan to work on the most concrete activities first, such as designing levels, and work our way up to the more abstract and blue-sky activities like mechanic prototyping. At the end of the section we discuss some possible future directions for the creative activity modules.

\subsection{Level Design}
Level design currently assumes the existence of a ruleset (in order to playtest a level), although this need not necessarily be the case in the future, as a designer might experiment with shapes or ideas and put them to one side for use in a game, or to inspire a ruleset. For now, though, the system takes a ruleset and uses generative techniques to draw out levels and then test them for playability. The current implementation of ANGELINA uses an evolutionary system that places down primitive shapes, with some light weighting for symmetry (inspired by interviews we conducted with game designer Alan Hazelden). These levels are then evaluated by using MCTS agents to play the game without a priori knowledge of the ruleset.

This activity requires further development, to build richer evaluation systems (such as different types of MCTS agent, as well as different agent models to gauge curiosity or less directed play) and more varied generation approaches. Having a mix of approaches that the system can choose between increases the breadth of the creative decisions being made, and allows evaluations to consider level designs from different angles. This activity is the most complete in ANGELINA currently -- the game \textit{Before Venturing Forth}, which appears in illustrations and code listings in this paper, had its levels designed by ANGELINA. 

\subsection{Ruleset Design}
Ruleset design uses a catalogue of known rule patterns to generate lists of trigger events that correspond to rules. A rule pattern is not necessarily captured by a single trigger. For example, the rule \textit{Lock And Key} might have two triggers: one for collecting the key, and one for touching doors after the key has been collected. Rulesets are created by selecting a number of rules from the catalogue, and then tested using simple level templates. These templates are designed to see if any kind of potential exists within the ruleset, rather than testing for depth or excitement or other more desirable qualities. For example, we might use an empty room with one of each game object around the edges, and run MCTS and random agents to see what triggers are activated, and whether the game ever ends. 

Initially this rule pattern catalogue is populated with known game mechanics from well-known genres and games. This is similar to the setup for many previous iterations of ANGELINA, although the rules are broader and more flexible than the later versions of ANGELINA which were quite restricted in the nature of the objectives they could offer. However, over time this catalogue will be extended to include rule patterns that ANGELINA has devised from its experimentation in the Mechanic Invention activity, outlined below.

\subsection{Mechanic Invention}
Mechanic invention refers to the creation of minor rule patterns which can be later used in the design of rulesets. This may involve one or more triggers which work together to create an effect in the game world. Our methodology here mirrors our earlier work on Mechanic Miner, an offshoot of ANGELINA which was never integrated into the main automated game design system \cite{mechanicminer}. Mechanic Miner assessed invented game mechanics by including them in an environment where certain things were known to be unachievable (such as reaching a certain point in the level) and then running exhaustive playtesters on the game. If the addition of the new mechanic now made some of these situations achievable, the new mechanic had clearly affected the player's possibilities in some way, and was thus deemed interesting.

We plan to expand this notion to cover more situations (accessibility was a good metric, but is quite specific and leads to a certain kind of mechanical discovery) and have multiple sample environments to test mechanics on, as well as more things to look out for. Our hope is not just that ANGELINA will find new game mechanics - but that it will find new mechanics without any specific idea or intention for them, and only later find that they suit a particular situation or inspire a particular kind of game design. By isolating activities, we allow for more freeform creativity, and put fewer demands on the output of the system.

\begin{figure}[h]
\centering
\includegraphics[width=0.8\columnwidth]{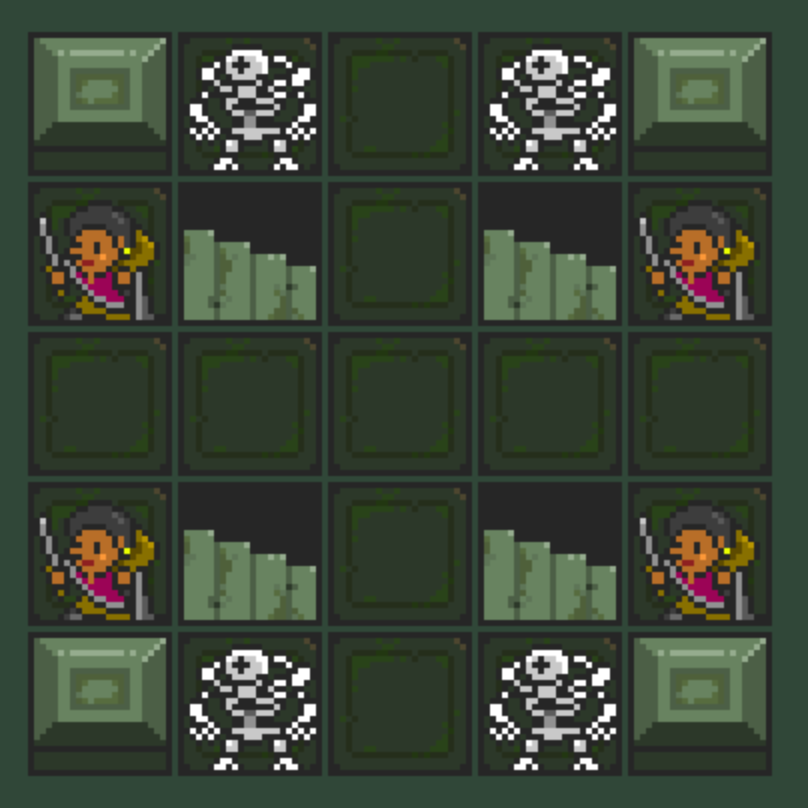}\\
\vspace{4mm}
\includegraphics[width=0.8\columnwidth]{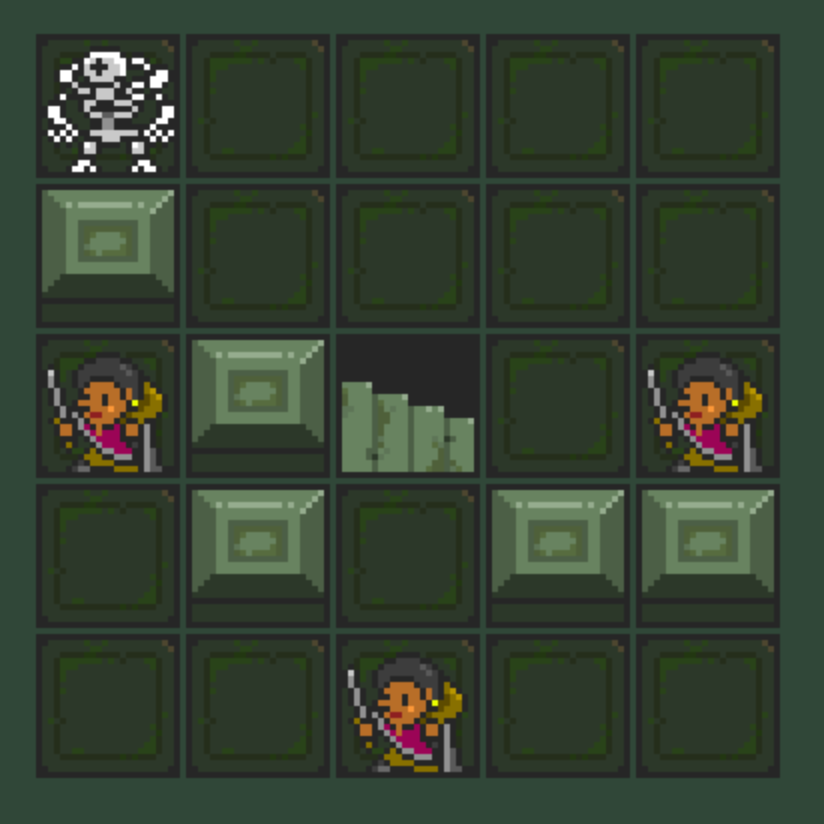}
\caption{Two levels designed for \textit{Before Venturing Forth} by ANGELINA. One level shows a more symmetric approach, while the other is more organic.}
\label{levels}
\end{figure}

\subsection{Milestone Case Study: Before Venturing Forth}
As we mentioned above, \textit{Before Venturing Forth} is a small game developed using ANGELINA's Level Design module and our prototype MCTS solving system. Figure \ref{levels} shows two levels designed by ANGELINA for the game. The symmetric influence in the level designer is very obvious in one level, whereas another leans more towards an asymmetric, organic design. 

It's interesting to note that even at this early stage, we are seeing emergence in the way ANGELINA is designing levels and exploring rulesets. \textit{Before Venturing Forth} was based on the \textit{Adventure} game template in the VGDL library, where the player controls an adventurer who defeats monsters and searches for an exit. However, with no constraint for the number of player characters, ANGELINA designed a game in which the objective is to get \textit{at least one} adventurer out alive, meaning the solution often involves intentionally sacrificing some of your team in order to get the rest out safely. While this is serendipitous at the moment, we're encouraged by these unusual findings and hope to get such interesting results intentionally recognised by the system in future.

Our MCTS agents are extremely basic currently - they attempt to play the game, with a small increased reward for finding new game states, but with no other metric for detecting distance to goal (since we are agnostic to the ruleset before beginning play). As ANGELINA develops, we plan to incorporate analysis of the game tree each MCTS bot creates, to understand the properties of the game being played (something which has previously been used to analyse the quality of board games\footnote{This work was conducted by Cameron Browne et al, although we are unable to locate citations for the work at the time of writing.}). For now this is strictly hypothetical, and our main priority is expanding the system vertically by including more activities for it to work with and shift between.

\section{Future Work}
The activities outlined above represent the first milestone for this new project, that allows the core of the system to begin doing creative work within game design. There are also plans for other activities that perhaps represent longer-term goals for when this system is active and developing. Growing systems through adding activities is as important to the development of automated game designers as the original system structure, so it's important to also study and think about how to extend existing systems with new functionality that seamlessly integrate with existing creative practices.

One such activity is \textit{theme exploration}. In this paper the notion of surface-level theming is left largely unexplored. We believe that exploring conceptual knowledge, talking to people online and noting down ideas for interesting themes and relationships could be built into a single creative activity designed to capture the feeling of exploring ideas and knowledge about the world. The output of such an activity might be stored maps that suggest ideas for games or visual mappings (not unlike the concept maps that humans would send as input to the Game-o-Matic). These might sit dormant and not be used in games for a long time; two might be combined together in a moment of inspirational discovery. This would pull in previous work on recovering information from online databases and seeking knowledge directly through social media \cite{phdthesis}.

Another activity we are considering is \textit{playing games}. Previously, it's not been possible for any version of ANGELINA to play games designed by people, because it used a closed engine and an obscure data format, meaning that not only did no humans design using the format, but very few `classic' games could be defined in it anyway. However, this new file format is far more open than before, and far closer to commonly-used standards like PuzzleScript. If we were to build a tool that allowed for simple design of these games, ANGELINA could conceivably have a database of human-designed games that it could play and critique. This would be an important activity not only in terms of potentially gaining new ideas for rulesets or level designs, but also for engaging with a creative community -- ANGELINA could provide feedback, reference other people in its work, or recommend games to other people.

Finally, we hope to extend ANGELINA's abstract rule invention further than simply defining new rules using existing keywords. In the engine that runs ANGELINA's games, keywords like OVERLAPS and DESTROY are interpreted at runtime and other code is executed to carry out these effects. Ultimately, we hope it would be possible for ANGELINA to invent new keywords in this language, by generating code and evaluating it in a similar way to Mechanic Miner or the current Mechanic Invention activity. This would be a much harder task, and take a long time to verify anything or find results. But it would show that ANGELINA was able to work at the highest and lowest levels of abstraction, and give even more breadth to its creative work.

\section{Conclusions}
In this paper we outlined our plan for a new structure for automated game designers, which breaks up the design process into smaller creative activities. These activities are designed to be performed at any time, in any order, as part of a \textit{continuous creative process} with no particular beginning or end. The creation of games becomes a natural consequence of creatively experimenting with game design, rather than the output of a big machine that is occasionally prodded into acting.

Our ultimate goal is to build a system which is fully exposed to people, so that they experience and perceive its creativity not simply through its outputs, but through the execution of the system itself. We are hoping to have ANGELINA create games over weeks, rather than hours, and to constantly post updates and work-in-progress samples through social media and other outlets. We are designing ANGELINA with the hope that the work that it does might be displayable visually (possibly at reduced speed) so that we could broadcast some or all of its creative process live on services like YouTube or Twitch (something that is not uncommon for other creative people to do\footnote{www.twitch.tv/directory/game/Creative}). We hope that by slowing the creative process down, emphasising independent work, and underlining the continuous and autonomous nature of the system, that we can raise the system's profile and have it accepted as a creative entity more than it ever has been in the past. And we believe many of these changes will also improve the quality of the games it makes, as a result of having a richer system of creation and growth.

We believe that this structural overhaul is vital in changing how we think about automated game design: from systems which are run like ordinary content generators that generate on demand, to autonomous systems that are constantly growing and changing, engaging in different kinds of creative activity and built in a general way that can be extended and played with. Automated game design has so many paths to explore, from traditional creativity support tools through to autonomous digital auteurs. We must keep branching out to make sure we don't miss any new paths or developers along our journey.

\section{Acknowledgements}
Thanks to the very kind reviewers who provided a variety of constructive feedback. We did our best to extend the paper appropriately. Thanks also to everyone who has provided feedback about ANGELINA over the years -- all of these discussions have informed this new version, and I hope it'll provoke more discussions in the future.

\bibliographystyle{aaai}
\bibliography{biblio}

\begin{thebibliography}{}

\bibitem[\protect\citeauthoryear{Browne}{2008}]{brownethesis}
Browne, C.
\newblock 2008.
\newblock {\em Automated Generation and Evaluation of Recombination Games}.
\newblock Ph.D. Dissertation, Queensland University of Technology, Queensland,
  Australia.

\bibitem[\protect\citeauthoryear{Colton and Wiggins}{2012}]{frontiers}
Colton, S., and Wiggins, G.~A.
\newblock 2012.
\newblock Computational creativity: The final frontier?
\newblock In {\em Proceedings of the 20th European Conference on Artificial
  Intelligence}.

\bibitem[\protect\citeauthoryear{Colton}{2008}]{perception}
Colton, S.
\newblock 2008.
\newblock Creativity versus the perception of creativity in computational
  systems.
\newblock In {\em AAAI Spring Symposium: Creative Intelligent Systems}.

\bibitem[\protect\citeauthoryear{Cook \bgroup et al\mbox.\egroup
  }{2013}]{mechanicminer}
Cook, M.; Colton, S.; Raad, A.; and Gow, J.
\newblock 2013.
\newblock Mechanic miner: Reflection-driven game mechanic discovery and level
  design.
\newblock In {\em Proceedings of the EVOGames Workshop, Applications of
  Evolutionary Computation Conference}.

\bibitem[\protect\citeauthoryear{Cook, Colton, and
  Pease}{2012}]{angelina-aiide}
Cook, M.; Colton, S.; and Pease, A.
\newblock 2012.
\newblock Aesthetic considerations for automated platformer design.
\newblock In {\em Proceedings of the Conference on Artificial Intelligence in
  Interactive Digital Entertainment}.

\bibitem[\protect\citeauthoryear{Cook}{2015}]{phdthesis}
Cook, M.
\newblock 2015.
\newblock {\em Co-operative coevolution for computational creativity: a case
  study In videogame design}.
\newblock Ph.D. Dissertation, Imperial College, South Kensington, London, UK.

\bibitem[\protect\citeauthoryear{Lavelle}{2014}]{puzzlescript}
Lavelle, S.
\newblock 2014.
\newblock Puzzlescript.
\newblock http://www.puzzlescript.net/.

\bibitem[\protect\citeauthoryear{Liapis, Yannakakis, and
  Togelius}{2013}]{sentient}
Liapis, A.; Yannakakis, G.~N.; and Togelius, J.
\newblock 2013.
\newblock Sentient sketchbook: Computer-aided game level authoring.
\newblock In {\em Proceedings of ACM Conference on Foundations of Digital
  Games}.

\bibitem[\protect\citeauthoryear{Nelson and Mateas}{2008}]{meaning}
Nelson, M.~J., and Mateas, M.
\newblock 2008.
\newblock An interactive game-design assistant.
\newblock In {\em Proceedings of the 13th International Conference on
  Intelligent User Interfaces}.

\bibitem[\protect\citeauthoryear{Pell}{1992}]{metagame}
Pell, B.
\newblock 1992.
\newblock Metagame in symmetric, chess-like games.
\newblock In {\em Heuristic Programming in Artificial Intelligence 3: The Third
  Computer Olympiad}.

\bibitem[\protect\citeauthoryear{Schaul}{2013}]{schaul2013pyvgdl}
Schaul, T.
\newblock 2013.
\newblock A video game description language for model-based or interactive
  learning.
\newblock In {\em Proceedings of the IEEE Conference on Computational
  Intelligence in Games}.
\newblock IEEE Press.

\bibitem[\protect\citeauthoryear{Shaker, Shaker, and Togelius}{2013}]{ropossum}
Shaker, N.; Shaker, M.; and Togelius, J.
\newblock 2013.
\newblock Ropossum: An authoring tool for designing, optimizing and solving cut
  the rope levels.
\newblock In {\em Proceedings of the Ninth {AAAI} Conference on Artificial
  Intelligence and Interactive Digital Entertainment}.

\bibitem[\protect\citeauthoryear{Smith, Whitehead, and Mateas}{2010}]{tanagra}
Smith, G.; Whitehead, J.; and Mateas, M.
\newblock 2010.
\newblock Tanagra: a mixed-initiative level design tool.
\newblock In {\em Proceedings of the Foundations of Digital Games Conference}.

\bibitem[\protect\citeauthoryear{Treanor \bgroup et al\mbox.\egroup
  }{2012}]{gom}
Treanor, M.; Schweizer, B.; Bogost, I.; and Mateas, M.
\newblock 2012.
\newblock The micro-rhetorics of game-o-matic.
\newblock In {\em Proceedings of the Foundations of Digital Games Conference}.
\newblock ACM.

\end{thebibliography}

\end{document}